%% file: root.tex
\documentclass[letterpaper, 10 pt, conference]{ieeeconf}  

\IEEEoverridecommandlockouts                              

\usepackage{graphicx}
\usepackage{caption}
\usepackage{subcaption}
\let\labelindent\relax
\usepackage{enumitem}
\usepackage{adjustbox}
\usepackage{multirow}
\usepackage{xcolor}
\usepackage{hyperref}
\usepackage{amsmath}

\newcommand{\namemh}{{\em NeuroLiDAR}}
\newcommand{\namemhs}{{\em NeuroLiDAR\ }}

\newcommand{\datasetmh}{{\em ELiDAR}}
\newcommand{\datasetmhs}{{\em ELiDAR\ }}

\title{\LARGE \bf
\namemh: Adaptive Frame Rate Depth Sensing via Neuromorphic Event-LiDAR Fusion
}

\author{Darshana Rathnayake$^{1}$, Dulanga Weerakoon$^{2}$, Meera Radhakrishnan$^{3}$ and Archan Misra$^{1}$
\thanks{$^{1}$Singapore Management University, Singapore, \newline 
{\tt\small darshanakg.2021@smu.edu.sg}, {\tt\small archanm@smu.edu.sg}}%
\thanks{$^{2}$Singapore-MIT Alliance for Research and Technology Centre, Singapore, 
{\tt\small dulanga.weerakoon@smart.mit.edu}}%
\thanks{$^{3}$University of Technology Sydney, Australia, \newline
{\tt\small meeralakshmi.radhakrishnan@uts.edu.au}}
}

\begin{document}

\maketitle
\thispagestyle{empty}
\pagestyle{empty}

\input{abstract}
\input{introduction}
\input{related-work}
\input{background}

\input{system-design}

\input{evaluation}
\input{discussion}

\section*{Acknowledgment}

This work was supported in part by: 1) National Research Foundation, Prime Minister’s Office, Singapore under its Campus for Research Excellence and Technological Enterprise (CREATE) program. The Mens, Manus, and Machina (M3S) is an interdisciplinary research group (IRG) of the Singapore-MIT Alliance for Research and Technology (SMART) centre; and 2) the Ministry of Education (MOE) Academic Research Fund (AcRF) Tier 2 grant (Grant ID: T2EP20124-0055). Any opinions, findings, and conclusions or recommendations expressed in this material are those of the author(s) and do not necessarily reflect the views of the granting agencies or the university.

\bibliographystyle{IEEEtran}
\bibliography{references}

\end{document}

%% file: abstract.tex
\begin{abstract}
LiDARs are widely used for 3D depth reconstruction, but their performance is often limited by inherent hardware constraints that impose trade-offs between range, spatial resolution, and frame rate.  Many LiDAR systems typically operate at low frame rates (e.g., 5-10 Hz), prioritizing long-range sensing over responsiveness to rapid scene changes. We present \namemh, an adaptive depth sensing framework that achieves effective frame rates of up to $\approx$66 Hz by fusing temporally sparse LiDAR data with temporally dense inputs from neuromorphic event cameras. \namemhs integrates two components: event-based keyframe detection and event-guided depth extrapolation, to dynamically adjust the sensing rate in response to scene dynamics. To evaluate our approach, we introduce \datasetmh, a dataset spanning outdoor and indoor scenarios, and show that \namemhs reduces depth reconstruction error by $\approx$29\% in RMSE while achieving adaptive frame rates between 27.8–47.3 Hz. Our code and dataset are available at \url{https://github.com/darshanakgr/neurolidar}.

\end{abstract}

%% file: introduction.tex
\section{Introduction}

LiDARs have become indispensable in automotive and industrial applications, such as autonomous driving and robotic navigation, due to their ability to provide accurate and reliable 3D depth measurements. Commercial automotive LiDARs typically operate at frame rates between 5-10 Hz, depending on the sensor’s range and resolution. While increasing the frame rate could improve the temporal resolution of depth tracking, it is known to both incur substantial energy overheads and reduce the operational lifespan of the laser source (i.e., the laser diode). 

This creates a fundamental tradeoff. In relatively static environments, operating a LiDAR sensor at a high frame rate consumes power unnecessarily without improving performance. In contrast, dynamic scenes with fast-moving objects or sudden vehicle manoeuvres require more frequent depth estimation than current LiDARs can provide. Ideally, a depth sensing system should operate at a low baseline LiDAR rate in static conditions, while adaptively increasing its effective frame rate when scene dynamics require it. Such an \emph{adaptive frame rate} paradigm could simultaneously prove to be both energy-efficient and responsive.

To achieve this dual objective and enable high-fidelity depth estimation without requiring the LiDAR to operate continuously at high frame rates, we propose to integrate a complementary sensing modality, \emph{neuromorphic event cameras}, that can capture environmental \emph{changes} at much lower system power ($\sim500$ mW) and finer temporal resolution ($O(1\mu s)$). Unlike conventional frame-based cameras, which capture images at fixed intervals, event cameras generate events asynchronously, at microsecond resolution~\cite{Gallego_2022}, whenever the incident light intensity changes principally due to motion dynamics. Our key idea is to operate the LiDAR itself to capture depth maps at a low base rate, while fusing this depth estimate with the motion dynamics captured continuously by the event camera to synthetically generate additional depth maps at appropriate intermediate time instants. 

Event cameras have been shown to complement RGB cameras in tasks such as image deblurring~\cite{lei2023}, depth densification~\cite{gehrig_combining_2021, cui2022dense}, video frame interpolation~\cite{yunfan2023}, and optical flow estimation~\cite{cuadrado2023}. Their asynchronous nature makes them particularly well suited to fill the temporal gaps between consecutive LiDAR frames in an \emph{adaptive} fashion, generating a higher volume of motion cues exactly when changes occur rapidly. Recent works have explored event–LiDAR fusion, primarily targeting depth densification--i.e., synthetically generating super-resolution depth estimates. For example, Li et al.~\cite{li2021enhancing} fused events with sparse LiDAR scans to produce up to $9.6\times$ denser point clouds, but with limited throughput ($\approx$4 fps). Other efforts have focused on filling sparse long-range LiDAR depth maps~\cite{cui2022dense, brebion2023learning}. Our work is, however, the first to propose \namemh, a novel depth sensing framework that \emph{performs adaptive \textbf{temporal} super-resolution, increasing the effective depth sensing frame rate by leveraging the asynchronous, low-latency dynamics captured by an event camera}, without incurring the energy cost of high frame-rate LiDAR scanning.

Unlike prior spatial densification methods, \namemhs performs adaptive depth extrapolation, producing new depth frames \emph{only} when events signal substantial changes to the scene. \namemhs operates in two stages: (a) a lightweight event-driven \emph{Keyframe Detector} identifies significant scene changes relative to the most recent LiDAR depth frame (prior depth frame) and then triggers the generation of synthetic depth estimates at those key moments, while (b) a U-Net-style~\cite{ronneberger2015u} lightweight autoencoder leverages the prior (most recently captured) reference depth frame, with a voxel grid-based event representation, to generate these depth estimates via extrapolation. This closed-loop design allows \namemhs to deliver high temporal fidelity only when required, while exploiting LiDAR’s inherently high spatial sensing resolution. We make the following \textbf{key contributions}:
\begin{itemize}[leftmargin=*]
   \item We present \emph{\textbf{Neuro}}morphic \emph{\textbf{LiDAR}} (\namemh), a novel depth sensing system that boosts LiDAR’s frame rate through \emph{event-guided depth extrapolation}, going beyond prior densification approaches. Our system also enables \emph{adaptive frame-rate LiDAR operation}, whereby the depth frames are generated, in a streaming fashion, only when scene dynamics warrant such generation, providing high temporal resolution without higher LiDAR scan rates or power costs.
   \item We built a real prototype of \namemh, carefully architected for high frame rate operation on embedded devices, using a commercial event camera and a LiDAR platform. We demonstrate its low-latency operation on resource-constrained embedded devices: when deployed on an NVIDIA Jetson Orin device, \namemhs can support an effective frame rate up to 66.67 Hz.
   \item To rigorously evaluate the performance of \namemh, we curate a new benchmark dataset, \datasetmh, which consists of both a large-scale synthetic data segment generated using the CARLA simulator~\cite{dosovitskiy2017carla} and a smaller real-world segment collected with our \namemhs prototype. The synthetic segment provides synchronized depth and event data under significantly more realistic and diverse driving conditions (e.g., varying speeds, traffic densities, pedestrian activities, abrupt, and changing environmental settings), compared to prior work such as ALED~\cite{brebion2023learning}. Experiments on the synthetic segment show that \namemhs achieves $\approx$29\% reduction in depth estimation error (RMSE) compared to a conventional LiDAR operating at 10 Hz. Subsequently, by experimentally deploying the \namemhs prototype on a mobile robot, we demonstrate that \namemhs can achieve a similar $\approx$28\% reduction in depth estimation RMSE in an indoor setting.
\end{itemize}

%% file: related-work.tex
\section{Related Work}

We discuss prior works on event-based depth sensing as well as event-based fusion approaches.

\noindent \textbf{Event-Only Depth}: Event cameras have found use in motion-centric tasks such as optical flow~\cite{cuadrado2023}, gesture recognition~\cite{chen2020dynamic}, action recognition~\cite{liu2021event}, SLAM~\cite{kim2016real, guan2024evi}, and video super-resolution~\cite{yunfan2023}, largely due to their ability to capture fast dynamics with minimal motion blur. This same property makes them attractive for depth estimation in dynamic scenes. Spiking neural networks (SNNs) have been leveraged for stereo event pairs~\cite{wu2024event}, which improved accuracy on MVSEC dataset~\cite{zhu2018multivehicle} by $\sim$20\% and offered 58x energy efficiency relative to ANNs. Earlier Kim et al.~\cite{kim2016real} used probabilistic filters for simultaneous localization and mapping (SLAM), though the reconstructions were semi-sparse and evaluated qualitatively. More recent systems such as EVI-SAM~\cite{guan2024evi} perform robust visual–inertial fusion with dense mapping, but at modest mapping frequencies ($\sim$7 Hz). While demonstrating the potential of purely event-based depth sensing, these works fall short of providing dense, accurate, and high frame-rate depth reconstructions.

\noindent \textbf{Event–RGB Fusion for Depth Estimation}: Early efforts to integrate event data with RGB cameras primarily focused on monocular depth prediction. Gehrig et al.~\cite{gehrig_combining_2021} extended recurrent neural architectures to jointly process irregular event streams and RGB frames, yielding up to 30\% lower mean absolute depth error than frame-based baselines. Other methods leveraged event–RGB fusion under adverse conditions: EVEN~\cite{shi2023even} enhanced RGB images with GANs before fusing them with events for robust depth estimation at night. More recently, hybrid spiking-transformer architectures~\cite{tumpa2024snn} combined event-driven SNN feature extraction with RGB-based transformers for depth prediction, though the latency remained high ($\approx$1 s per frame on GPUs, 171 ms on custom hardware). These works demonstrated the complementary value of events and frames, but remain limited by the range and passive stereo constraints of RGB inputs.

\noindent \textbf{Event–LiDAR Fusion for Depth Densification}: Recent works have addressed densifying sparse LiDAR depth maps using event streams. Li et al.~\cite{li2021enhancing} estimated depth for event pixels using neighbouring LiDAR points, yielding $\approx$9.6× denser point clouds and improved detection accuracy, but limited to semi-dense reconstructions at $\approx$4 fps.  More recently, multi-stage approaches~\cite{cui2022dense, brebion2023learning} have been explored for spatial depth map densification using LiDAR and event data. While the work~\cite{cui2022dense} reported throughput values up to 56 fps, the definition of latency was unclear, and the evaluation of reconstruction accuracy was limited to depth values of 50 m. Collectively, these methods improve the spatial resolution of LiDAR but do not target temporal adaptivity.

Our proposed \namemhs framework is thus motivated by the unresolved problem of adaptively increasing LiDAR’s effective frame rate under fast, real-world dynamics, spanning both outdoor and indoor scenarios. 

%% file: background.tex
\section{Background and Motivation}

Reliable depth perception in dynamic environments, such as encountered in autonomous driving or robot navigation applications, requires sensors that can capture rapid scene changes. However, current commercial LiDARs remain bound by low frame rates, leaving critical gaps that demand new sensing strategies, as well as dedicated datasets to evaluate such strategies.

\begin{figure}
    \centering
    \includegraphics[width=0.99\linewidth]{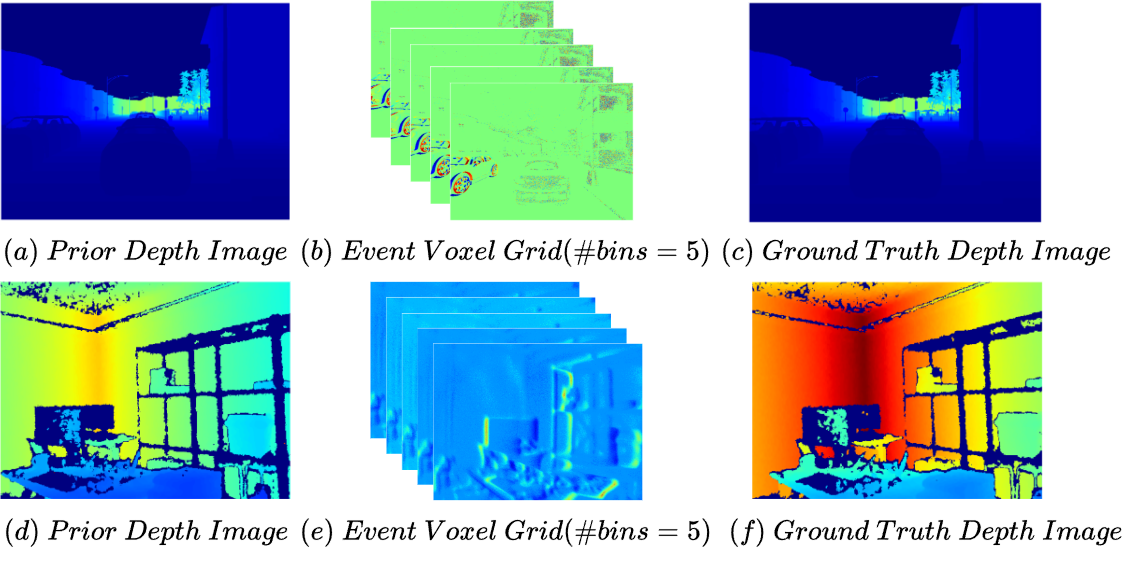}
    \caption{Examples from the \datasetmhs dataset: (a)–(c) simulated outdoor scenarios, (d)–(f) real-world indoor scenarios.}
    \label{fig:dataset}
    \vspace{-0.205in}
\end{figure}

\subsection{Construction of \datasetmhs Dataset}

For training and evaluation of \namemh, we construct a dataset termed \datasetmh. Our dataset contains two segments for a simulated outdoor scenario and a real-world indoor scenario. The simulated segment is generated using the CARLA simulator \cite{dosovitskiy2017carla, Hidalgo20threedv}, which captures diverse urban driving scenarios with aligned LiDAR and event camera streams, recorded at 200 Hz and configured for long-range sensing (up to 200 m). Both the LiDAR and event cameras are mounted on the roof of the ego vehicle within CARLA and configured to capture at a spatial resolution of $480 \times 640$ pixels. The fields of view of the two sensors are aligned to ensure pixel-level correspondence between the depth and event modalities. The simulated split of the dataset spans varied lighting, weather, traffic, and pedestrian conditions, with ego vehicle speeds ranging from slow cruising to highway motion. This configuration allows us to generate a large scale, temporally dense dataset for training and evaluating both the keyframe detection and depth estimation models. 

The real-world segment of \datasetmhs is collected with our \namemhs prototype, mounted on a robotic platform which traversed different trajectories, at varying speeds, within an indoor room environment. The LiDAR was configured to operate at 30 Hz, the maximum frame rate supported by the Intel RealSense L515. This relatively small split helps assess the generalizability of \namemhs from simulated outdoor scenarios to real-world indoor settings.

Figure~\ref{fig:dataset} shows representative samples from the \datasetmhs dataset, including scenes captured under both simulated outdoor and real-world indoor settings. In total, simulated splits of \datasetmhs contain 244 separate 20s long sequences, captured under 12 predefined weather conditions. The real-world indoor split contains $15$ additional 10s long sequences. \datasetmhs fills a key gap in the availability of real-world high-frame-rate LiDAR data and provides a benchmark for both keyframe detection and adaptive depth estimation.

\subsection{Motivation for a High Frame Rate LiDAR}

\begin{figure}[!htb]
    \centering
    \begin{subfigure}[t]{0.49\linewidth}
        \centering
        \includegraphics[width=\linewidth]{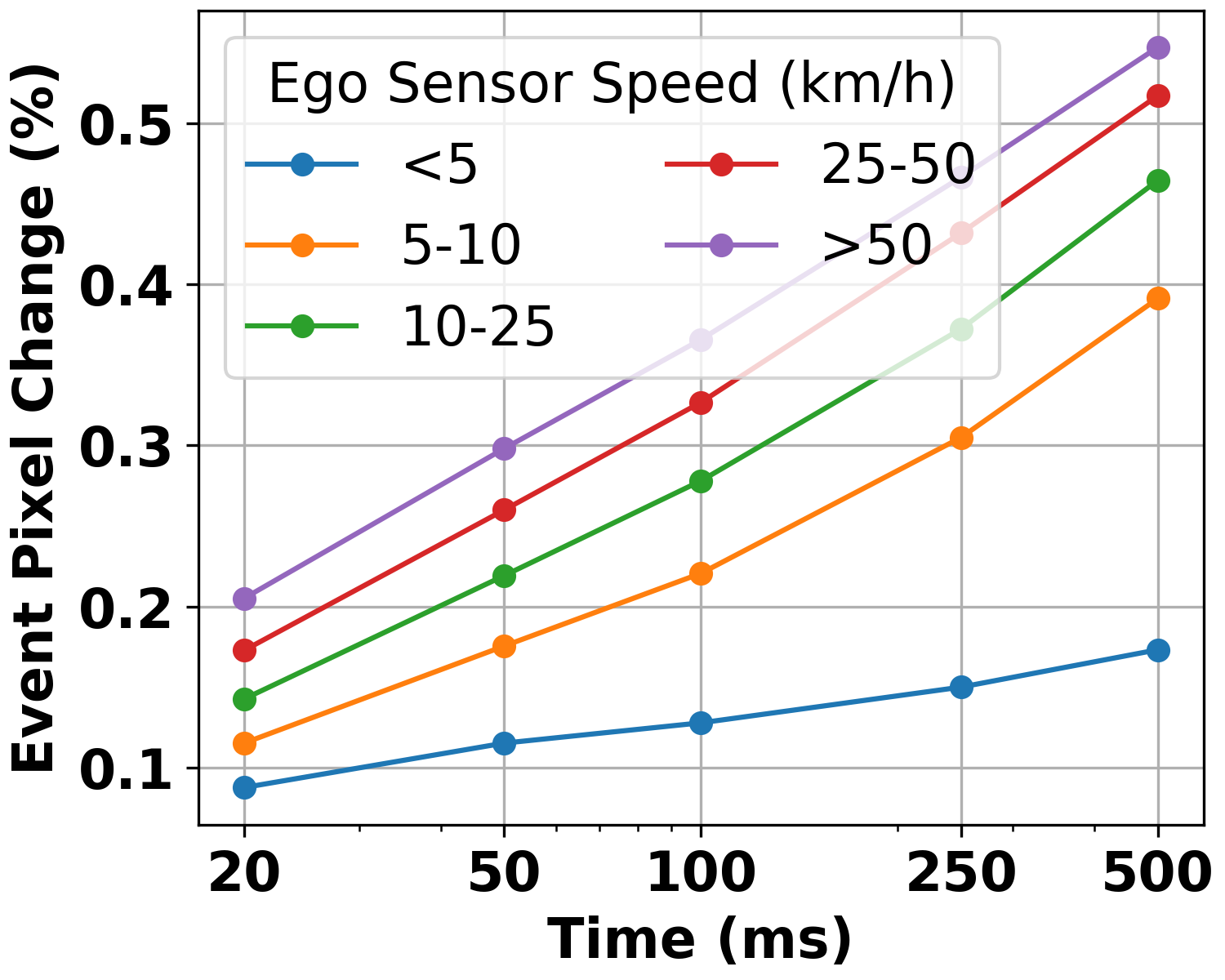}
        \caption{Event change over time}
        \label{fig:event_change}
    \end{subfigure}
    \hfill
    \begin{subfigure}[t]{0.49\linewidth}
        \centering
        \includegraphics[width=\linewidth]{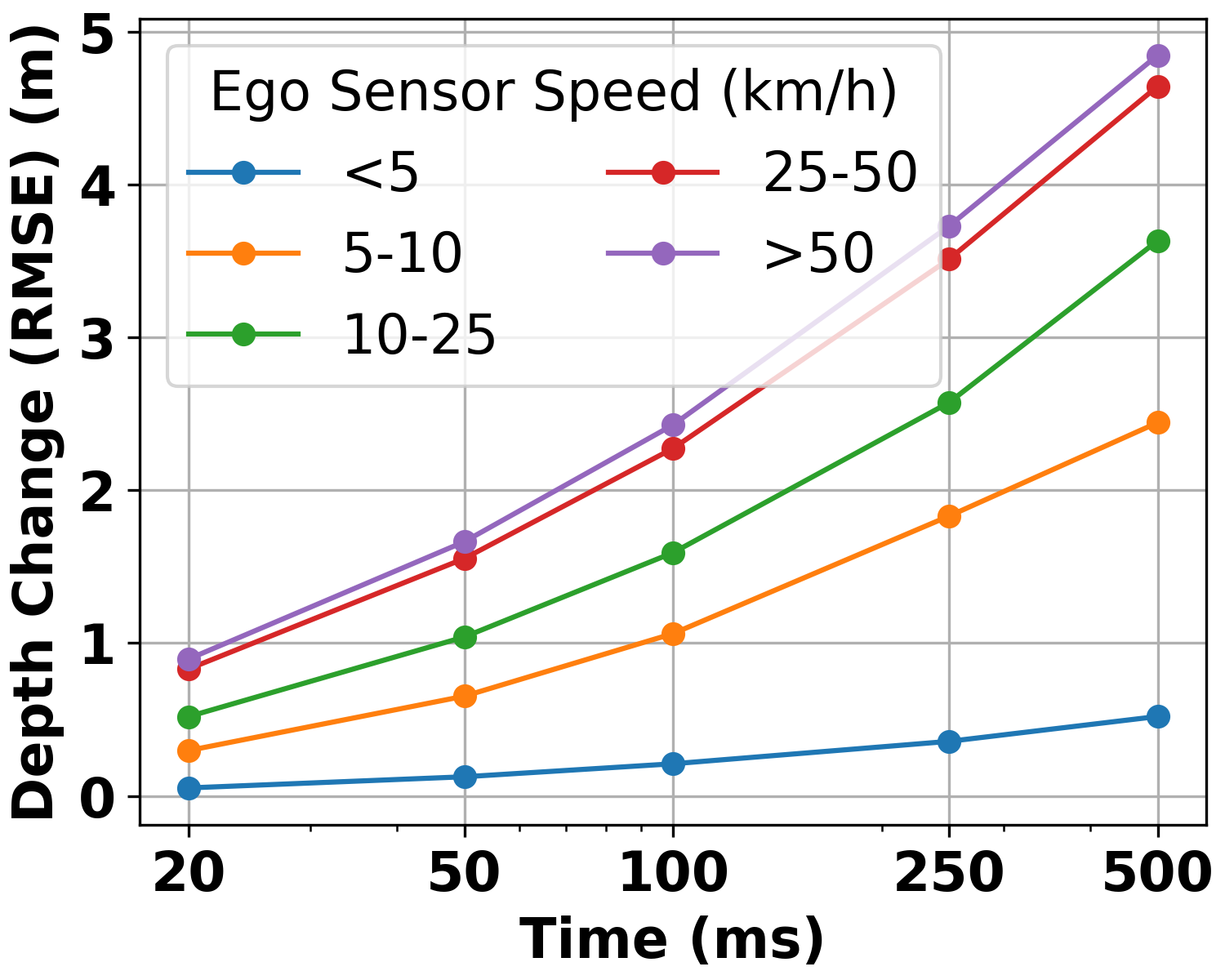}
        \caption{Depth change over time}
        \label{fig:depth_change}
    \end{subfigure}
    \caption{Event and depth changes with time for varying ego-sensor speeds}
    \label{fig:preliminary_eval}
\end{figure}

Commercial automotive LiDARs typically operate at frame rates between 5-10Hz, depending on the sensor’s range and resolution. In principle, their frame rate can be increased, but this comes with trade-offs: higher frame rates reduce effective range and resolution, substantially increase power consumption, and may shorten the lifespan of the sensor due to limitations of the laser diode. These limitations render current commercial LiDARs inadequate for reliably capturing rapid scene changes in highly dynamic environments.

Analysis of \datasetmhs highlights the temporal limitations of current LiDAR systems. We measured the change in depth over different time intervals and vehicle velocities (Figure \ref{fig:depth_change}). For a 100 ms interval (corresponding to 10 Hz, a common frame rate for outdoor LiDARs such as Velodyne~\cite{velodyne}), we observed an average depth change of approximately $2.4$ m when the vehicle was moving at $\geq$ 50 km/h, a typical driving speed. Such changes exceed the spatial accuracy needed for reliable perception, revealing that these LiDARs are insufficient for fast-moving scenarios.

\subsection{Motivation for an Adaptive Frame Rate LiDAR}

While we have established the need to expand LiDAR capabilities to support higher frame rates, not all driving conditions require high temporal fidelity. For instance, as shown in Figure \ref{fig:depth_change}, when the ego vehicle moves at $\leq$ 10 km/h, the average change in depth is $\leq$ 1 m, making high-frequency depth sampling unnecessary (compared to when the vehicle travels at $\geq$ 50 km/h). A fixed high frame rate would therefore waste energy, highlighting the need for an \emph{adaptive frame rate LiDAR} that responds to scene dynamics.

Achieving adaptivity requires a lightweight sensing mechanism to infer scene changes without activating the high-power LiDAR sensor. High frame rate RGB cameras could serve this purpose, but their power demands\footnote{https://www.shodensha.co.th/product/183316-173555/chu30} ($\approx$3.3\,W) make them impractical for embedded deployment. In contrast, neuromorphic event cameras offer ultra-high temporal resolution ($\approx$1\,MHz) at low power ($\approx$0.5\,W). By asynchronously capturing pixel-level brightness changes, event cameras are particularly well-suited to directly capture motion-driven scene changes with low latency, high sensitivity, and low energy overheads. Such detected changes can, in turn, trigger explicit depth sensing by an adaptive frame rate LiDAR.

We conducted a preliminary analysis on \datasetmhs to evaluate whether event sensor outputs can effectively capture scene dynamics (Figure \ref{fig:event_change}). Specifically, we vary the ego vehicle’s speed and compute the fraction of active event pixels relative to the total ($640 \times 480$) pixel count. Across increasing time windows and vehicle speeds, we consistently observe that the proportion of active event pixels rises, mimicking the trend for depth changes reported in Figure \ref{fig:depth_change}. \emph{This demonstrates that event representations can reliably capture motion dynamics in the scene and exhibit strong correlation with depth changes, making them well-suited as a triggering mechanism for adaptive depth estimation.}

%% file: system-design.tex
\section{\namemhs System Design}
\label{sec:system_design}

We now outline the design goals, high-level architecture, and implementation of our \namemhs system.

\subsection{Design Goals}
\begin{itemize}[leftmargin=*]
\item \textbf{Adaptive Frame Rate:} The system should dynamically adjust its frame rate in response to perceived changes in scene dynamics. In particular, \namemhs must support a maximum effective frame rate that exceeds the native frame rate of the baseline LiDAR sensor.

\item \textbf{Low Computational Complexity:} To sustain high effective frame rates when needed, the depth estimation process must be lightweight enough for real-time execution on an embedded device, such as the Jetson platform. Specifically, the inference latency $l$ must satisfy $l \leq 1/F$, where $F$ is \namemh's current operating frame rate.

\item \textbf{Reduced Depth Extrapolation Error:} Despite being lightweight, the model should minimize depth extrapolation error across a range of frame rates, ensuring robustness without sacrificing accuracy.
\end{itemize}

\subsection{System Architecture}

\begin{figure}[t]
    \centering
    \begin{minipage}[c]{0.55\columnwidth}
        \centering
        \includegraphics[width=\linewidth,height=1.7in]{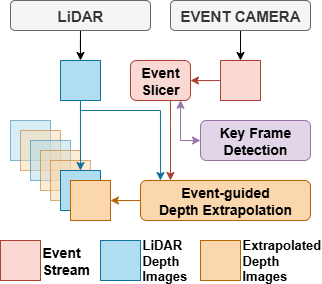}
        \caption{High-Level Architecture}
        \label{fig:architecture}
    \end{minipage}
    \hfill
    \begin{minipage}[c]{0.43\columnwidth}
        \centering
        \includegraphics[width=\linewidth,height=1.7in]{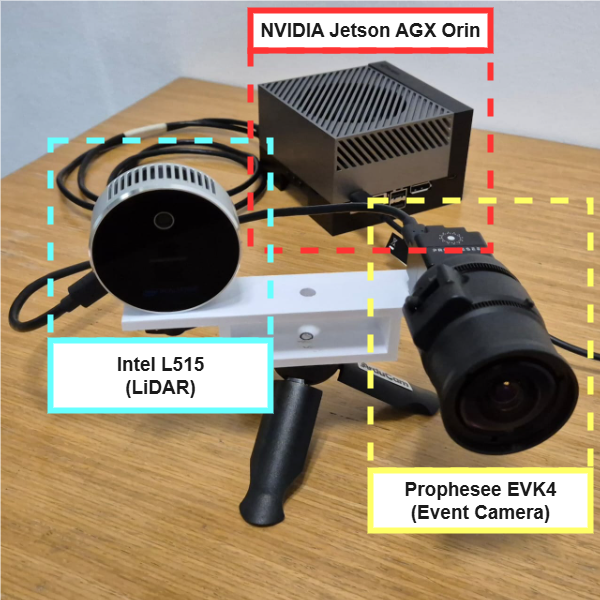}
        \caption{\namemh's System Implementation}
        \label{fig:neurolidar}
    \end{minipage}
    \vspace{-0.1in}
\end{figure}

We develop \namemhs by coupling an event camera with a base, low-frame-rate LiDAR. The event stream captured by the neuromorphic camera serves two complementary roles: (a) identifying the moments when new depth frames should be generated (i.e., the reference LiDAR depth frames should be augmented via extrapolation), and (b) providing motion priors for the actual depth extrapolation process. Figure~\ref{fig:architecture} presents the high-level system architecture, comprising three key components:

\begin{enumerate}[leftmargin=*]
\item \textbf{Event Slicer:} Processes the raw event stream into structured event representation(s) that can be used by both the keyframe detection and depth extrapolation modules.
\item \textbf{Keyframe Detection:} Determines whether an event slice corresponds to a significant scene change and, if so, triggers the depth extrapolation module via Event Slicer.
\item \textbf{Event-guided Depth Extrapolation:} Combines the most recent LiDAR depth image with accumulated events since its capture to generate an extrapolated depth image.
\end{enumerate}

Keyframe detection and depth extrapolation are performed by two separate DNN models, both running concurrently on a representative low-power device (Jetson AGX Orin), resulting in an adaptive frame rate \namemh. In addition, to balance accuracy vs. computational efficiency, \namemhs uses two distinct event representations: (i) the computationally cheaper, less descriptive event frame representation for the keyframe detector, and (ii) the more descriptive voxel grid representation~\cite{gallego2020event} for Event-guided Depth Extrapolation.

\subsection{System Implementation}

We implement the \namemhs prototype (illustrated in Figure~\ref{fig:neurolidar}) using an Intel RealSense L515 \cite{intelL515specs} as the low-frame-rate LiDAR sensor and a Prophesee EVK4 \cite{prophesee} as the event camera. The keyframe detection and depth extrapolation models are deployed on an NVIDIA Jetson AGX Orin \cite{jetson}. To support real-time operation with reduced energy consumption, both models are quantized to 16-bit FP precision and executed using TensorRT.

\section{Adaptive Event-Guided Depth Extrapolation}

We now detail \namemh’s key functions. For a formal description, let the depth frame captured at time $t$ be denoted as $D_t \in \mathcal{R}^{H \times W}$, where $H$ and $W$ represent the height and width of the depth map, respectively. We define an asynchronous event stream within a time interval of $(t, t+\Delta)$ as $\mathcal{E}_{(t, t+\Delta)} = \{e_i = (p_i, x_i, y_i, t_i)\}_{i=0}^N$ where $p_i \in \{-1,1\}$ indicates the polarity corresponding to a brightness increase or decrease, $\{x_i, y_i\} \in \mathcal{R}^{H \times W}$ denotes the pixel coordinates, and $t_i$ denotes the timestamp.

\subsection{Event Slicer}

The event slicer accumulates asynchronously arriving events and converts them into structured event representations. Concretely, for an asynchronous event stream $\mathcal{E}_{(t, t+\Delta)}$, an event frame $\mathcal{E}^F(x,y)$ (consumed by the Event Slicer) is defined as follows:
\begin{equation}
    \mathcal{E}^F(x,y) = \sum \limits_{e_i\in \mathcal{E}_{(t,t+\Delta)}}p_i \mathbf{1}\{x=x_i,y=y_i\}
    \label{eq:frame}
\end{equation}

In our design, we set $\Delta = 20$ms and pass this \emph{event frame} representation to the keyframe detection model, which processes the frames and triggers the depth extrapolator whenever it identifies a \emph{significant change} in the scene. 

Assuming that a keyframe is detected at $t=t+t_1$, we then obtain an event voxel $\mathcal{E}^V(b,x,y)$ (consumed by the Event-guided Depth Extrapolation component) where $b\in [0,B)$ for $B=5$ as follows.

\begin{equation}
\resizebox{\columnwidth}{!}{$
    \mathcal{E}^V(b,x,y) = \sum \limits_{e_i\in \mathcal{E}_{(t,t+t_1)}}p_i 
    \mathbf{1}\{b = \left\lfloor B \cdot \frac{t_i - t}{t_1} \right\rfloor, x=x_i,y=y_i\}
$}
\label{eq:voxel}
\end{equation}

\begin{figure*}
    \centering
    \includegraphics[width=0.99\linewidth]{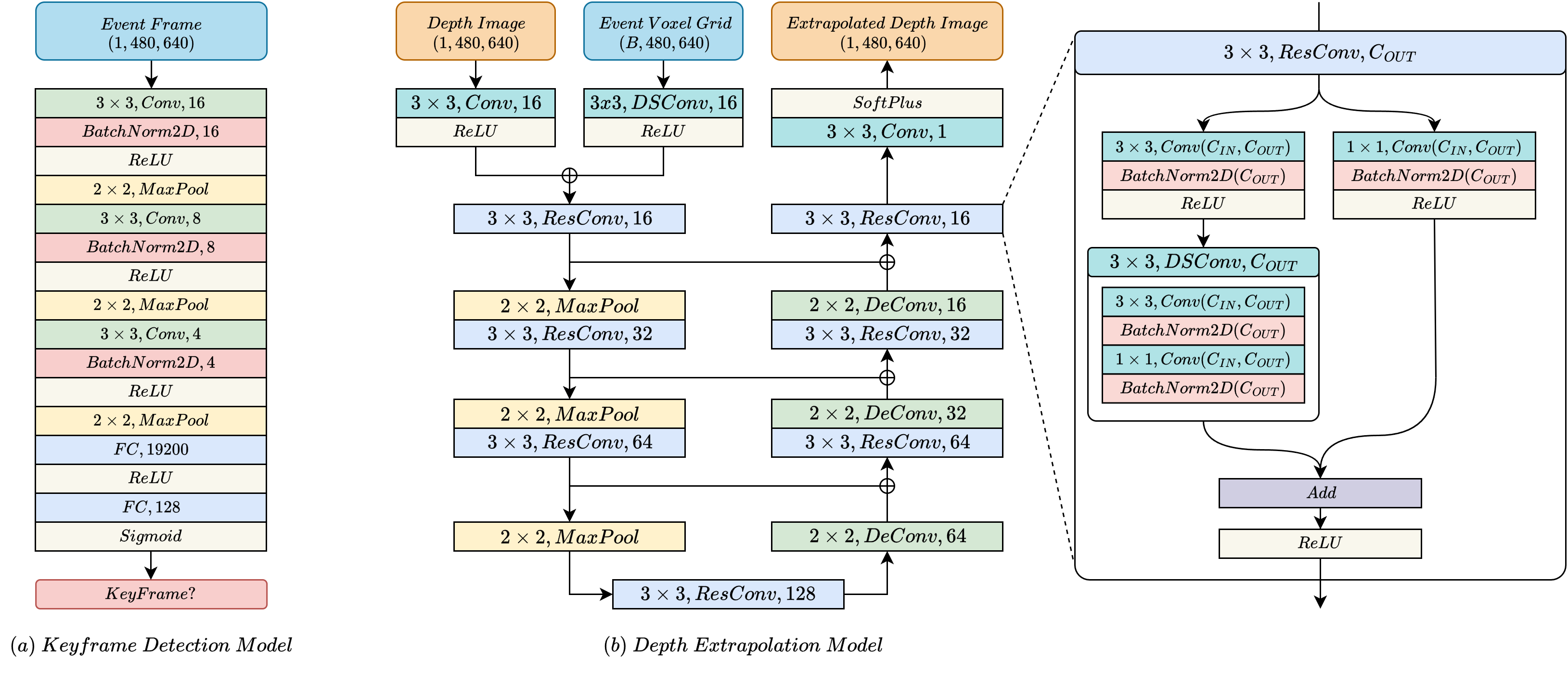}
    \caption{Neural network architectures of keyframe detection and depth extrapolation network with expanded view of ResConv}
    \label{fig:main}
    \vspace{-0.2in}
\end{figure*}

\subsection{Keyframe Detection Model}

A core feature of \namemhs is its ability to adaptively regulate frame rate based on scene dynamics, delivering high temporal fidelity in fast-changing environments while conserving energy when the environment is largely static. To enable this, we design a lightweight keyframe detection model that monitors incoming event streams and identifies moments of significant change, such as ego-motion or the appearance of moving objects (e.g., vehicles, pedestrians). Only when such keyframes are detected is the depth extrapolation module triggered, ensuring that new depth frames are generated precisely when needed rather than periodically.

\subsubsection{\textbf{Characterizing a Keyframe}} We define a keyframe as one that captures a significant scene change, typically arising in outdoor driving from dynamic objects such as vehicles and pedestrians and in indoor environments from unexpected motion of human occupants and other objects (e.g., robots, machinery). To operationalize this, we extract contextual cues from event sub-streams and use them to train our DNN model. We employ a threshold-based strategy guided by three indicators: ego-sensor speed, distance to surrounding objects, and the appearance of new objects in the field of view. For ground truth labelling, a keyframe is defined when any of the following conditions are satisfied:

\begin{itemize}[wide=0pt,topsep=0pt]
\item The ego sensor speed exceeds 10 m/s (=36 km/h).
\item The shortest distance to a surrounding object (such as a vehicle or pedestrian) falls below 8 meters.
\item A new object enters the sensor’s field of view.
\end{itemize}

\subsubsection{\textbf{Architecture of the Keyframe Detection Model}} We formulate keyframe detection as a binary classification problem, where each event frame is classified as either a keyframe or a non keyframe. We design a lightweight three-layer convolutional neural network (illustrated in Figure~\ref{fig:main}(a)) that takes an event frame defined in equation \ref{eq:frame} as input and outputs a binary classification result. The model is trained using the binary cross entropy loss function for $20$ epochs with the NAdam optimizer, batch size of $64$, and a learning rate of $1 \times 10^{-4}$. The dataset contains 24,720 training and 11,220 test samples, randomly drawn from 114 training and 50 test sequences in the simulated outdoor segment.

\subsection{Event-guided Depth Extrapolation}

Once the keyframe detector predicts a significant change in the scene at time $t=t_1$, we then activate the event-guided depth extrapolation model. This model takes two inputs: (a) a prior depth image ($D_t$), which is the most recent depth image captured by the LiDAR, and (b) the event voxel representation from Event Slicer (Equation~\ref{eq:voxel}) in the time interval $(t,t_1)$. The events provide motion information relative to the prior depth image and guide the extrapolation task. To satisfy the real-time requirements of our system, we design the model architecture to be lightweight and consistent with the keyframe detection model, ensuring that each depth frame is predicted before the arrival of the next frame. The model is a streamlined U-Net architecture, illustrated in Figure \ref{fig:main}(b), composed of convolution, max-pooling, and transposed convolution operations followed by ReLU activations. The final output layer, which generates the depth image, applies a SoftPlus activation. The architecture follows an encoder–decoder structure that employs max-pooling and transposed convolution to downsample and upsample the spatial dimensions of the feature maps. The intermediate convolutional layers consist of two parallel convolution operations (3x3 and 1x1): the 1×1 convolution layer extracts features from the input in parallel to 3×3 convolution, and its output feature maps are added to those of the depthwise separable convolution before the final activation.

For training, we adopt a combination of loss functions $\left(\mathcal{L}_{depth}, \mathcal{L}_{grad}, \mathcal{L}_{norm}, \mathcal{L}_{SSIM}\right)$, employed in METER \cite{papa2023meter}. For $\mathcal{L}_{depth}$, we use MSE loss instead of L1, with weighting factors $\lambda_2=10, \lambda_3=0.01, \lambda_4=1$. Our model is trained for $50$ epochs using the NAdam optimizer with an initial learning rate of $1\times10^{-3}$ and a cosine annealing scheduler. The training dataset consists of 22,543 training samples and 5,597 test samples, randomly drawn from 80 training sequences and 16 test sequences within the simulated outdoor segment.

%% file: evaluation.tex
\section{Results}

We now evaluate the \namemh's performance on both the simulated outdoor and real indoor segments of \datasetmh. 

\subsection{Evaluation Metrics}

    Following prior work on depth estimation \cite{godard2019digging}, we adopt standard depth evaluation metrics to benchmark the performance of \namemh. In addition to accuracy based measures, we also evaluate system level efficiency by reporting the inference latency and, retrospectively, the maximum effective frame rate that \namemhs can achieve on an NVIDIA Jetson AGX Orin. To benchmark depth extrapolation accuracy, we use the following standard metrics:

\begin{enumerate}[leftmargin=*]
\item RMSE: $\sqrt{\frac{1}{|D|}\sum_{d \in D} (d - \hat{d})^2}$
\item log RMSE: $\sqrt{\frac{1}{|D|}\sum_{d \in D} (\log d - \log \hat{d})^2}$
\item AbsRel: $\frac{1}{|D|}\sum_{d \in D} \frac{|d - \hat{d}|}{d}$
\item SQRel: $\frac{1}{|D|}\sum_{d \in D} \frac{(d - \hat{d})^2}{d}$
\item $\delta$: \% of $\hat{d}$ s.t. $\max\left(\frac{d}{\hat{d}}, \frac{\hat{d}}{d}\right) < \delta$, for $\delta \in {1.25, 1.25^2, 1.25^3}$
\end{enumerate}

Additionally, we adopt standard classification metrics: Precision, Recall, and F1-score, to evaluate the efficacy of the keyframe detection model. Unless specified otherwise, all evaluations of \namemhs are conducted on the synthetic outdoor driving segment of \datasetmh.

\begin{figure}[htbp]
    \centering
    \begin{subfigure}[t]{0.48\linewidth}
        \centering
        \includegraphics[width=\linewidth]{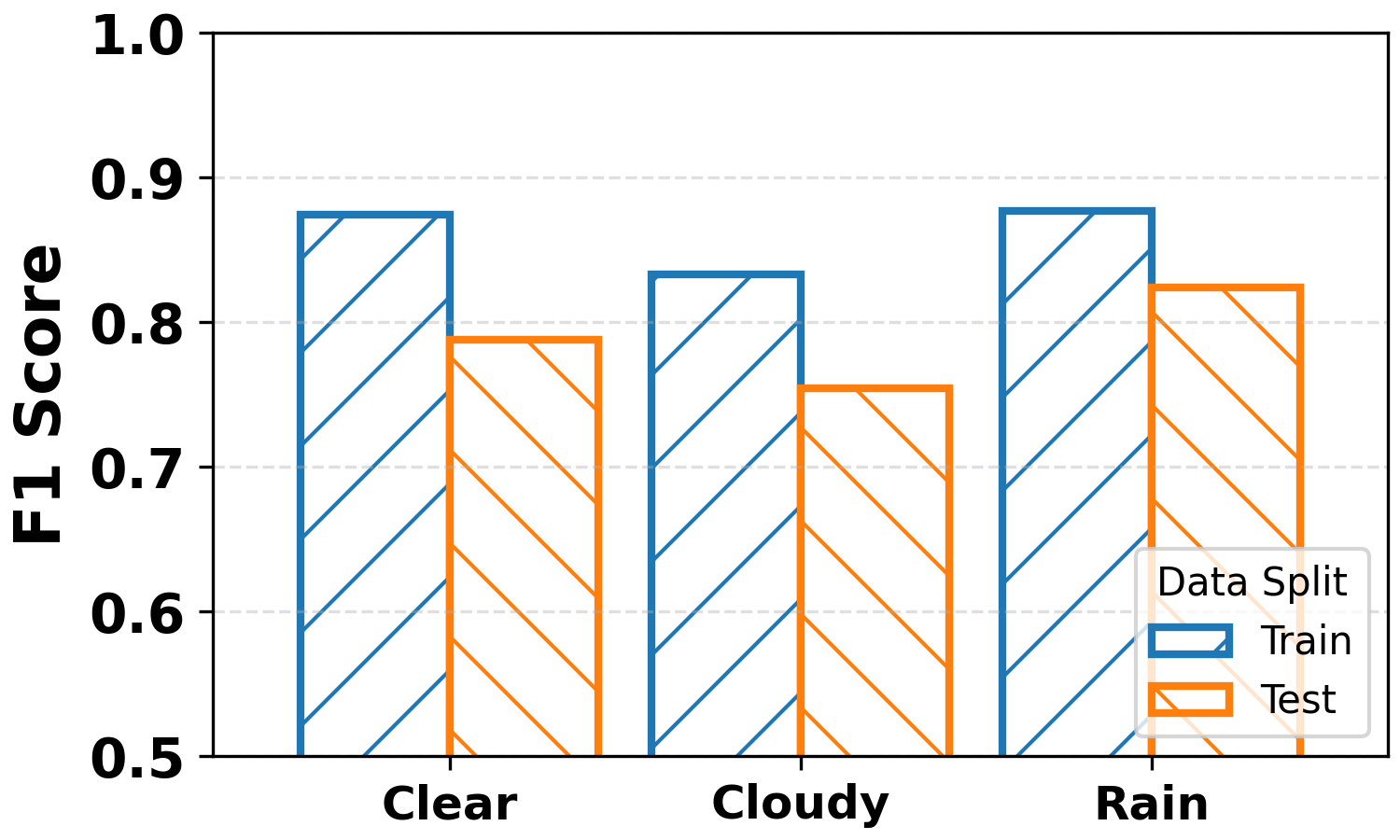}
        \caption{Different weather conditions}
        \label{fig:weather_key}
    \end{subfigure}
    \hfill
    \begin{subfigure}[t]{0.48\linewidth}
        \centering
        \includegraphics[width=\linewidth]{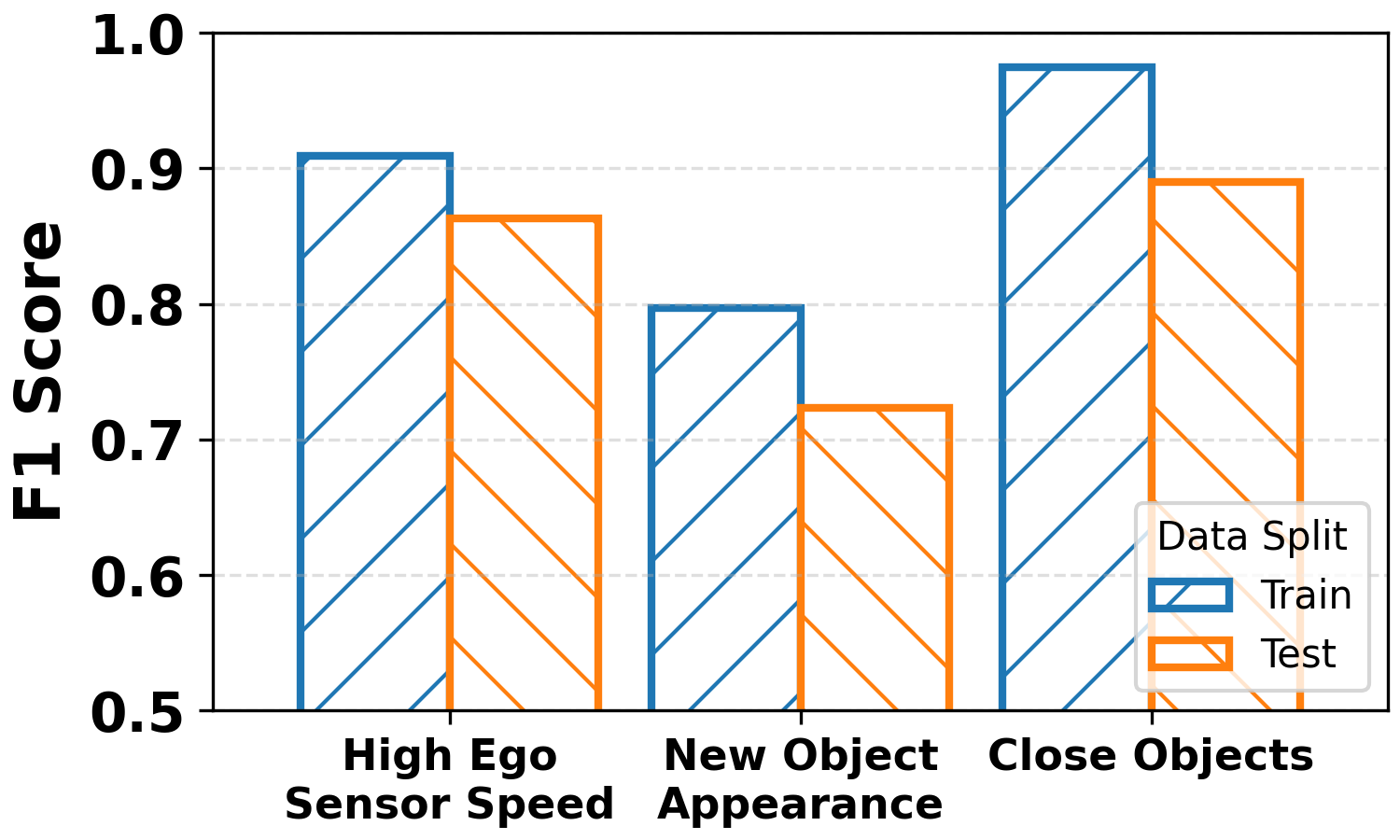}
        \caption{Keyframe characteristics}
        \label{fig:motion_key}
    \end{subfigure}

    \caption{Keyframe detection under different conditions}
    \label{fig:motion_weatherKey}
    \vspace{-0.15in}
\end{figure}

\subsection{Performance Evaluation of Keyframe Detection Model}

\begin{table}[]
\caption{\namemh's keyframe detection accuracy (\datasetmhs Simulated)}
\label{tab:keyframe_TrainTestAcc}
\centering
\begin{tabular}{l|l|l|l}
\hline
\textbf{Split} & \textbf{F1-Score} & \textbf{Precision} & \textbf{Recall} \\ \hline
\hline
Train & 0.888    & 0.893     & 0.884  \\ 
Test  & 0.798    & 0.825     & 0.773  \\ \hline
\end{tabular}
\vspace{-0.12in}
\end{table}

Table \ref{tab:keyframe_TrainTestAcc} summarizes the performance of the keyframe detection model across the simulated segment of \datasetmhs dataset. On the test dataset, our model achieves an F1-score of 79.8\%. The temporal gap between a false positive and either a true positive or the next actual LiDAR depth image is observed to be low, varying between 41.7--88.6 ms, underscoring that the keyframe detection module can reliably identify the optimal moments to trigger depth estimation (with a fairly tight error bound) while remaining lightweight and computationally efficient. Importantly, this efficiency ensures that extrapolated depth frames can be produced in real time, thereby maintaining the high and adaptive frame rates required to handle fast and dynamic motion sequences.

In Figures \ref{fig:weather_key} and \ref{fig:motion_key}, we further analyze the F1-score of keyframe detection under different weather conditions and motion/object dynamics. Our model achieves strong performance in detecting high-speed motion and nearby objects, reaching F1-scores of 86\% and 89\% respectively. The performance slightly decreases when identifying newly appearing objects (F1-score=72\%). This reduction may stem from the insufficient temporal span across frames. Nevertheless, the lightweight keyframe detector consistently delivers F1-scores between 75.4\% and 82.8\% across diverse weather conditions, demonstrating both robustness and effectiveness.

\subsection{Performance Evaluation of Depth Extrapolation Model}

\begin{table}[]
\caption{Depth estimation error for different frame rates: \namemhs vs. other baselines}
\label{tab:acc_fps}
\scriptsize
\setlength{\tabcolsep}{4pt} 
\resizebox{\columnwidth}{!}{
\begin{tabular}{c|c|c|c|c|c|c|c|c}
\hline
\textbf{Method}           & \textbf{\begin{tabular}[c]{@{}c@{}}FPS\\$(F)$\end{tabular}} & \textbf{RMSE $\downarrow$} & \textbf{\begin{tabular}[c]{@{}c@{}}Log\\ RMSE $\downarrow$\end{tabular}} & \textbf{\begin{tabular}[c]{@{}c@{}}ABS \\ Rel $\downarrow$\end{tabular}} & \textbf{\begin{tabular}[c]{@{}c@{}}SQ \\ Rel $\downarrow$\end{tabular}} & \textbf{\begin{tabular}[c]{@{}c@{}}$\boldsymbol{\delta}$\textless\\ 1.250 $\uparrow$\end{tabular}} & \textbf{\begin{tabular}[c]{@{}c@{}}$\boldsymbol{\delta}$\textless\\ 1.562 $\uparrow$\end{tabular}} & \textbf{\begin{tabular}[c]{@{}c@{}}$\boldsymbol{\delta}$\textless\\ 1.953 $\uparrow$\end{tabular}} \\ \hline
\hline
\multirow{6}{*}{Repeat} & 2            & 12.211        & 3.487                                                       & 0.139                                                       & 4.521                                                      & 0.860                                                               & 0.911                                                               & 0.933                                                               \\
                          & 5            & 9.820         & 2.980                                                       & 0.092                                                       & 2.603                                                      & 0.910                                                               & 0.940                                                               & 0.954                                                               \\
                          & 10           & 8.382         & 2.373                                                       & 0.063                                                       & 1.884                                                      & 0.941                                                               & 0.960                                                               & 0.970                                                               \\
                          & 20           & 6.904         & 2.016                                                       & \textbf{0.045}                                                       & 1.341                                                      & 0.958                                                               & 0.971                                                               & 0.978                                                               \\
                          & 50           & 4.976         & 1.486                                                       & \textbf{0.026}                                                       & 0.697                                                      & \textbf{0.975}                                                               & 0.984                                                               & 0.987                                                               \\  
                          & Adap.        & 8.414         & 2.455                                                       & 0.072                                                       & 2.187                                                      & 0.929                                                               & 0.953                                                               & 0.964                                                               \\ \hline \hline
\multirow{6}{*}{Ours}     & 2            & \textbf{8.422}         & \textbf{0.528}                                                       & \textbf{0.109}                                                       & \textbf{2.075}                                                      & \textbf{0.885}                                                               & \textbf{0.944}                                                               & \textbf{0.967}                                                               \\
                          & 5            & \textbf{6.385}         & \textbf{0.394}                                                       & \textbf{0.074}                                                       & \textbf{1.010}                                                      & \textbf{0.929}                                                               & \textbf{0.967}                                                               & \textbf{0.981}                                                               \\
                          & 10           & \textbf{5.542}         & \textbf{0.292}                                                       & \textbf{0.055}                                                       & \textbf{0.689}                                                      & \textbf{0.953}                                                               & \textbf{0.979}                                                               & \textbf{0.988}                                                               \\
                          & 20           & \textbf{4.747}         & \textbf{0.237}                                                       & 0.046                                                       & \textbf{0.500}                                                      & \textbf{0.963}                                                               & \textbf{0.984}                                                               & \textbf{0.991}                                                               \\
                          & 50           & \textbf{3.883}         & \textbf{0.185}                                                       & 0.036                                                       & \textbf{0.325}                                                      & 0.974                                                               & \textbf{0.989}                                                               & \textbf{0.994}                                                               \\ 
                          & Adap.        & \textbf{5.769}         & \textbf{0.325}                                                       & \textbf{0.063}                                                       & \textbf{0.910}                                                      & \textbf{0.941}                                                               & \textbf{0.972}                                                               & \textbf{0.984}                                                               \\ \hline
\end{tabular}
}
\end{table}

We compare baseline extrapolation methods with our event-guided depth extrapolation model in Table~\ref{tab:acc_fps}. The effective frame rate $F$ of \namemhs is also shown, where extrapolation raises the base LiDAR rate from $F/2$ to $F$. To ensure fair evaluation under fixed conditions, we test across fixed frame rates ranging from 2–50 Hz. In the adaptive (\emph{Adap.}) setting, we randomly sample frames from the dataset at varying frame rates, with the model again extrapolating a single intermediate frame. For both fixed and adaptive settings, we compare our depth extrapolation model against a \textit{Repeat} baseline, which simply reuses the sampled depth frame at $F/2$ as the extrapolated frame. Across all metrics, our event-guided extrapolation consistently outperforms this baseline, demonstrating robustness under both static and adaptive frame rate conditions, e.g., comparing \emph{Adap.} setting, our approach has 31.43\% lower RMSE than \textit{Repeat}.

\subsection{Ablation Studies for Event-Guided Depth Extrapolation}

\begin{table}[t]
\centering
\caption{Ablation results on input type, model variants, and loss functions.}
\label{tab:ablation}
\scriptsize
\setlength{\tabcolsep}{3pt} 
\resizebox{\columnwidth}{!}{
\begin{tabular}{p{1.9cm}|c|c|c|c|c|c|c}
\hline
\multicolumn{1}{c|}{\textbf{Method}} & 
\multicolumn{1}{c|}{\textbf{RMSE $\downarrow$}} & 
\multicolumn{1}{c|}{\textbf{\begin{tabular}[c]{@{}c@{}}Log\\ RMSE $\downarrow$\end{tabular}}} & 
\multicolumn{1}{c|}{\textbf{\begin{tabular}[c]{@{}c@{}}ABS \\ Rel $\downarrow$\end{tabular}}} & 
\multicolumn{1}{c|}{\textbf{\begin{tabular}[c]{@{}c@{}}SQ \\ Rel $\downarrow$\end{tabular}}} & 
\multicolumn{1}{c|}{\textbf{\begin{tabular}[c]{@{}c@{}}$\boldsymbol{\delta}$\textless \\ $\mathbf{1.25}\uparrow$\end{tabular}}} &
\multicolumn{1}{c|}{\textbf{\begin{tabular}[c]{@{}c@{}}$\boldsymbol{\delta}$\textless \\ $\mathbf{1.56}\uparrow$\end{tabular}}} &
\multicolumn{1}{c}{\textbf{\begin{tabular}[c]{@{}c@{}}$\boldsymbol{\delta}$\textless \\ $\mathbf{1.95\uparrow}$\end{tabular}}} \\
\hline \hline
\multicolumn{8}{c}{\textbf{Input-based Ablations}} \\ 
w/o Event         & 7.799 & 0.364 & 0.103 & 1.693 & 0.911 & 0.952 & 0.967 \\
Depth + Ev. frames & 6.161    & 0.332   & 0.071   & 1.082   & 0.933   & 0.968   & 0.981   \\ \hline \hline
\multicolumn{8}{c}{\textbf{Model Ablations}} \\ 
Data concat        & 5.936   & \textbf{0.315}   & 0.073   & 0.954   & 0.936   & 0.970   & 0.983   \\
w/o skip           & 5.935   & 0.392   & 0.079   & 0.985   & 0.931   & 0.969   & 0.983   \\ \hline \hline
Ours               & \textbf{5.769} & 0.325 & \textbf{0.063} & \textbf{0.910} & \textbf{0.941} & \textbf{0.972} & \textbf{0.984} \\ \hline
\end{tabular}
}
\end{table}

Table~\ref{tab:ablation} presents a detailed ablation study of \namemh's depth extrapolation model, considering two dimensions: (a) input-based and (b) model-based. Below, we describe the variants and their effects.

\begin{itemize}[wide=0pt,topsep=0pt]
\item \textbf{w/o Event:} The event encoder is removed, leaving the model without motion priors for extrapolation. This leads to a substantial degradation in performance across all metrics, increasing RMSE by +2.03 compared to the full model.
\item \textbf{Depth + Ev. frames:} Instead of voxel-grid representation, we use event frames (the same representation employed in the keyframe detection model). While this representation provides a coarse motion prior, it lacks the fine-grained temporal resolution of voxel grids, resulting in moderately higher errors, with RMSE increasing by +0.392.
\item \textbf{Data concat:} Our primary model encodes events and depth inputs separately, followed by feature-level fusion. This variant concatenates the events and depth as a single input and passes it through a single encoder. This results in a performance drop (RMSE +0.167), showing the importance of specialized encoders for different modalities.
\item \textbf{w/o skip:} This variant measures the impact of $1\times1$ convolution layer as a skip connection in ResConv Blocks. By removing it, the performance drops across all metrics, with RMSE increasing by +0.166, confirming their role in preserving spatial detail during depth extrapolation.
\end{itemize}

\subsection{End-to-End System Evaluation}

We now conduct an end-to-end system evaluation, including both the keyframe detection and depth extrapolation tasks on the test sequences of \datasetmh. In this experiment, we set the low baseline frame rate of LiDAR to 10 Hz. We first compare the performance against several baselines (reported in Table~\ref{tab:system_eval}):

\begin{itemize}[wide=0pt,topsep=0pt]
    \item \textbf{Repeat:} Most recent depth image is repeated as the extrapolated frame at the keyframe instance.
    \item \textbf{Linear:} Assuming the depth varies linearly over time, we regress the pixel-wise depth value temporally with a linear function ($d_{t+\Delta}=ad_{t} + b)$, where both $a$ and $b$ coefficients are determined by previous depth images
    \item \textbf{Exponential:} Similarly, we assume depth varies as an exponential function ($d_{t+\Delta} = ae^{\Delta d_{t}}$), where $a$ is estimated with the last two consecutive depth images.
\end{itemize}

When compared to these baselines, \namemhs reduces the RMSE by $\approx$1.81 (relative to the repeat baseline), achieving lower error across the evaluation metrics, while reaching a maximum frame rate of \textbf{47.30 Hz} (a \textbf{4.7$\boldsymbol{\times}$} increase over the standard 10 Hz LiDAR). On the \datasetmhs dataset, \namemhs adapts its frame rate between \textbf{27.81 Hz} and \textbf{47.30 Hz}, with a mean effective frame rate of \textbf{40.58 Hz}.

\begin{figure}[t]
    \centering
    \begin{minipage}[c]{0.48\columnwidth}
        \centering
        \includegraphics[width=0.98\linewidth,height=1.6in]{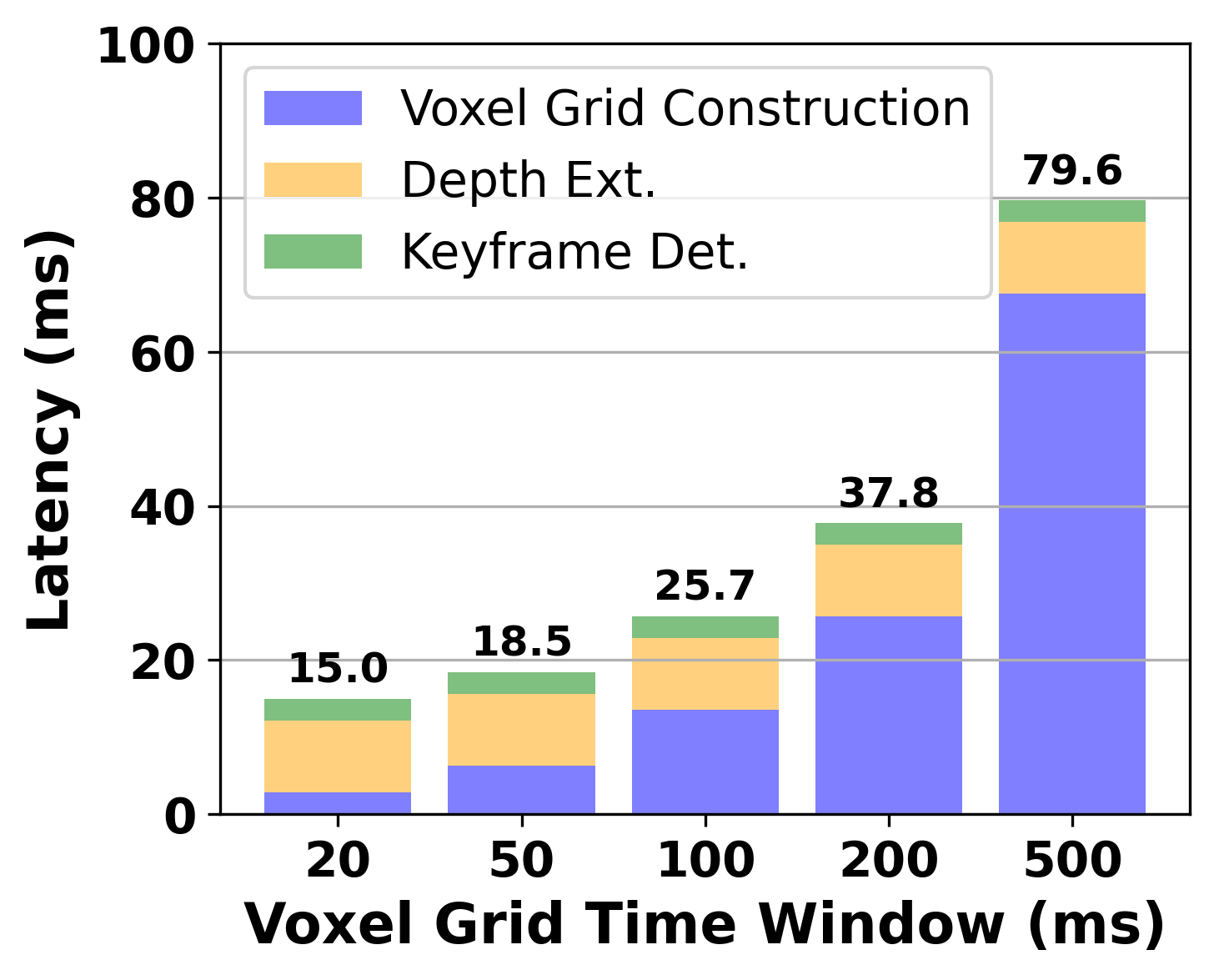}
        \caption{End-to-end latency including keyframe detection, voxel-grid construction, and depth extrapolation.}
        \label{fig:end-to-end-latency}
    \end{minipage}
    \hfill
    \begin{minipage}[c]{0.5\columnwidth}
        \centering
        \includegraphics[width=\linewidth,height=1.6in]{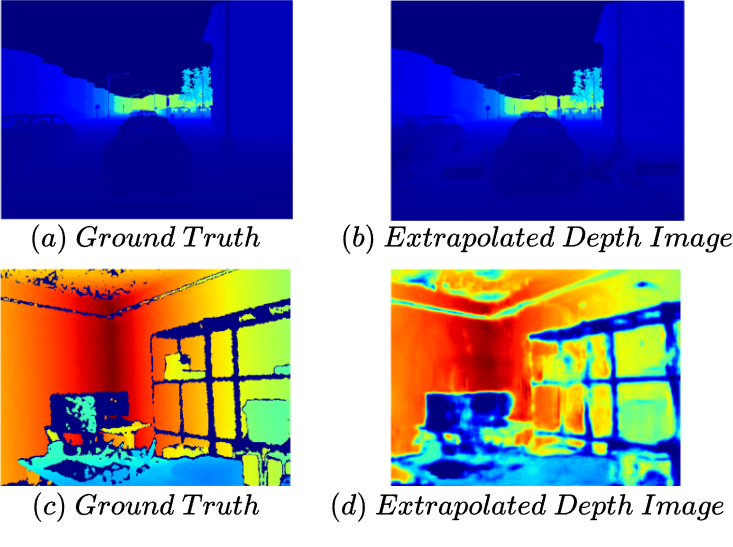}
        \caption{Ground truth vs. extrapolated depth images: (a–b) simulated outdoor, (c–d) real-world indoor.}
        \label{fig:predictions}
    \end{minipage}
    \vspace{-0.12in}
\end{figure}

In Figure \ref{fig:end-to-end-latency}, we further analyze the end-to-end latency of the system, which is determined by three main components: the event slicer, the keyframe detector, and the event-guided depth extrapolator. For a voxel-grid with a time window of 20 ms, \namemhs can theoretically operate within 15 ms (supporting up to 66.67 Hz), which is $\sim6.67\times$ the operating frequency of the LiDAR (10 Hz). While the latencies of the keyframe detector ($2.81$ ms) and the event-guided depth extrapolator ($9.31$ ms) remain constant across different voxel sizes, the latency of the event slicer depends on the event window size--i.e., the duration until the keyframe detector signals a significant scene change. A larger window size such as $500$\,ms means that the \namemhs needs to operate only at 2 Hz due to a largely static scene. Even then, end-to-end latency of the system is \textbf{79.6 ms}, which is far lower than the required 500 ms time window. \namemhs can thus operate at a lower effective frame rate in static scenes, thereby allowing more time for frame extrapolation.

\begin{table}[t]
\centering
\caption{End-to-end depth estimation comparison }
\label{tab:system_eval}
\scriptsize
\setlength{\tabcolsep}{3pt} 
\resizebox{\columnwidth}{!}{
\begin{tabular}{c|c|c|c|c|c|c|c}
\hline
\multicolumn{1}{c|}{\textbf{Method}} & 
\multicolumn{1}{c|}{\textbf{RMSE $\downarrow$}} & 
\multicolumn{1}{c|}{\textbf{\begin{tabular}[c]{@{}c@{}}Log\\ RMSE $\downarrow$\end{tabular}}} & 
\multicolumn{1}{c|}{\textbf{\begin{tabular}[c]{@{}c@{}}ABS \\ Rel $\downarrow$\end{tabular}}} & 
\multicolumn{1}{c|}{\textbf{\begin{tabular}[c]{@{}c@{}}SQ \\ Rel $\downarrow$\end{tabular}}} & 
\multicolumn{1}{c|}{\textbf{\begin{tabular}[c]{@{}c@{}}$\boldsymbol{\delta}$\textless \\ $\mathbf{1.25}\uparrow$\end{tabular}}} &
\multicolumn{1}{c|}{\textbf{\begin{tabular}[c]{@{}c@{}}$\boldsymbol{\delta}$\textless \\ $\mathbf{1.56}\uparrow$\end{tabular}}} &
\multicolumn{1}{c}{\textbf{\begin{tabular}[c]{@{}c@{}}$\boldsymbol{\delta}$\textless \\ $\mathbf{1.95\uparrow}$\end{tabular}}} \\
\hline \hline 
Repeat                & 6.226   & 1.912   & \textbf{0.045}   & 1.210   & 0.956   & 0.971   & 0.956   \\
Linear                & 8.248   & 2.758   & 0.075   & 1.992   & 0.926   & 0.952   & 0.963   \\
Exponential           & 25.131   & 3.589   & 0.454   & 10.683   & 0.255   & 0.520   & 0.622   \\
\namemhs               & \textbf{4.416} & \textbf{0.245} & 0.047 & \textbf{0.457} & \textbf{0.962} & \textbf{0.984} & \textbf{0.991} \\ \hline
\end{tabular}
}
\vspace{-0.12in}
\end{table}

\subsection{Generalizability}

In addition to evaluation on the synthetic outdoor \datasetmhs dataset, we demonstrate the generalizability of \namemhs for the real indoor dataset, collected using our prototype (Figure~\ref{fig:neurolidar}). Table~\ref{tab:eval_on_realworld} presents the resulting accuracy of \namemh's event-guided depth extrapolation model. For this, we finetuned the depth extrapolation model on the training split of the real-world subset of \datasetmhs for $50$ epochs with a learning rate of $1 \times 10^{-4}$ by freezing the encoder's parameters and scaling the sensor's range from $200$m to $9$m. 

Testing involved adaptively sampling depth frames at effective frame rates of up to 15 Hz and extrapolating a single intermediate frame from the test split of the real-world indoor dataset. Compared to the \emph{Repeat} baseline, \namemhs achieves substantial gains across most accuracy metrics, including a 28.61\% reduction in RMSE. While these results demonstrate promising generalization to indoor environments, additional large-scale evaluation in indoor environments such as warehouses and factory floors will be pursued as future work. We do not assess the generalizability of the keyframe detection module, as this scenario differs fundamentally from outdoor driving conditions.

\begin{table}[t]
\centering
\caption{\namemhs depth estimation error for the real-world indoor scenario}
\label{tab:eval_on_realworld}
\scriptsize
\setlength{\tabcolsep}{3pt} 
\begin{tabular}{c|c|c|c|c|c|c|c}
\hline
\multicolumn{1}{c|}{\textbf{Method}} & 
\multicolumn{1}{c|}{\textbf{RMSE $\downarrow$}} & 
\multicolumn{1}{c|}{\textbf{\begin{tabular}[c]{@{}c@{}}Log\\ RMSE $\downarrow$\end{tabular}}} & 
\multicolumn{1}{c|}{\textbf{\begin{tabular}[c]{@{}c@{}}ABS \\ Rel $\downarrow$\end{tabular}}} & 
\multicolumn{1}{c|}{\textbf{\begin{tabular}[c]{@{}c@{}}SQ \\ Rel $\downarrow$\end{tabular}}} & 
\multicolumn{1}{c|}{\textbf{\begin{tabular}[c]{@{}c@{}}$\boldsymbol{\delta}$\textless \\ $\mathbf{1.25}\uparrow$\end{tabular}}} &
\multicolumn{1}{c|}{\textbf{\begin{tabular}[c]{@{}c@{}}$\boldsymbol{\delta}$\textless \\ $\mathbf{1.56}\uparrow$\end{tabular}}} &
\multicolumn{1}{c}{\textbf{\begin{tabular}[c]{@{}c@{}}$\boldsymbol{\delta}$\textless \\ $\mathbf{1.95\uparrow}$\end{tabular}}} \\
\hline \hline 
Repeat                & 0.819   & 4.213   & 0.068   & 0.207   & \textbf{0.939}   & 0.942   & 0.944   \\
\namemhs               & \textbf{0.532} & \textbf{0.316} & \textbf{0.066} & \textbf{0.087} & 0.903 & \textbf{0.948} & \textbf{0.970} \\ \hline

\end{tabular}
\vspace{-0.12in}
\end{table}

%% file: discussion.tex
\section{Discussion}

\noindent \textbf{Increasing the Spatial Fidelity of LiDARs:} \namemhs has primarily focused on improving the temporal resolution of LiDAR sensing by adaptively extrapolating depth frames. While this effectively increases the frame rate up to 65+ Hz, the challenge of simultaneously enhancing spatial fidelity, i.e., increasing the point cloud resolution of LiDAR outputs, remains open. We speculate that the same DNN-based depth extrapolation framework presented here can be extended to address spatial resolution, applying techniques analogous to the use of U-Net-based autoencoders for event-guided depth densification~\cite{cui2022dense,brebion2023learning}. In future, we plan to explore how U-Net autoencoders can be adapted to jointly support both high temporal and spatial fidelity within a single network.

\noindent \textbf{Adaptive Bin Size for Event-Voxel Grids:} To further enhance the fidelity of spatiotemporal depth extrapolation, \namemhs can potentially utilize alternative non-uniform voxelization techniques, borrowed from point cloud compression literature, to create non-uniform event voxel grids. For example, the most recently constructed depth map may be used as a \emph{prior} to create finer voxels in regions characterized by sharp depth variations; such approaches may be especially useful for higher depth resolution required for fine-grained robot manipulation. However, such non-uniform voxelization requires further investigation for careful optimization, as it is likely to increase the processing latency of the \emph{Event Slicer}.

\section{Conclusion}

We presented \namemh, a novel LiDAR depth sensing system that combines a standard LiDAR with a neuromorphic event camera to achieve higher temporal fidelity through adaptive frame rates supporting up to 66.67 Hz.  \namemhs continuously uses the event stream to perceive scene changes and then triggers the depth sensing module to extrapolate depth frames at appropriate time instants. For depth extrapolation, we use a lightweight U-Net-style autoencoder to fuse a prior depth frame captured by the standard LiDAR with a motion representation from the high frame rate event camera. Evaluation performed using a new benchmark \datasetmhs dataset, which models both diverse urban driving scenarios and a smaller, real-world indoor environment, shows that  \namemhs reduces depth reconstruction error by $\approx$30\% compared to a standard LiDAR operating at 10 Hz.

%% file: root.bbl
\begin{thebibliography}{10}
\providecommand{\url}[1]{#1}
\csname url@rmstyle\endcsname
\providecommand{\newblock}{\relax}
\providecommand{\bibinfo}[2]{#2}
\providecommand\BIBentrySTDinterwordspacing{\spaceskip=0pt\relax}
\providecommand\BIBentryALTinterwordstretchfactor{4}
\providecommand\BIBentryALTinterwordspacing{\spaceskip=\fontdimen2\font plus
\BIBentryALTinterwordstretchfactor\fontdimen3\font minus \fontdimen4\font\relax}
\providecommand\BIBforeignlanguage[2]{{%
\expandafter\ifx\csname l@#1\endcsname\relax
\typeout{** WARNING: IEEEtran.bst: No hyphenation pattern has been}%
\typeout{** loaded for the language `#1'. Using the pattern for}%
\typeout{** the default language instead.}%
\else
\language=\csname l@#1\endcsname
\fi
#2}}

\bibitem{Gallego_2022}
G.~Gallego, T.~Delbruck, G.~Orchard, C.~Bartolozzi, B.~Taba, A.~Censi, S.~Leutenegger, A.~J. Davison, J.~Conradt, K.~Daniilidis, and D.~Scaramuzza, ``Event-based vision: A survey,'' \emph{IEEE Transactions on Pattern Analysis and Machine Intelligence}, vol.~44, no.~1, p. 154–180, 2022.

\bibitem{lei2023}
L.~Sun, C.~Sakaridis, J.~Liang, P.~Sun, K.~Zhang, J.~Cao, Q.~Jiang, K.~Wang, and L.~Van~Gool, ``Event-{Based} {Frame} {Interpolation} with {Ad}-hoc {Deblurring}.''\hskip 1em plus 0.5em minus 0.4em\relax IEEE, June 2023, pp. 18\,043--18\,052.

\bibitem{gehrig_combining_2021}
D.~Gehrig, M.~R{\"u}egg, M.~Gehrig, J.~Hidalgo-Carri{\'o}, and D.~Scaramuzza, ``Combining events and frames using recurrent asynchronous multimodal networks for monocular depth prediction,'' \emph{IEEE Robotics and Automation Letters}, vol.~6, no.~2, pp. 2822--2829, 2021.

\bibitem{cui2022dense}
M.~Cui, Y.~Zhu, Y.~Liu, Y.~Liu, G.~Chen, and K.~Huang, ``Dense depth-map estimation based on fusion of event camera and sparse lidar,'' \emph{IEEE Transactions on Instrumentation and Measurement}, vol.~71, pp. 1--11, 2022.

\bibitem{yunfan2023}
Y.~Lu, Z.~Wang, M.~Liu, H.~Wang, and L.~Wang, ``Learning spatial-temporal implicit neural representations for event-guided video super-resolution,'' in \emph{2023 IEEE/CVF Conference on Computer Vision and Pattern Recognition (CVPR)}, 2023, pp. 1557--1567.

\bibitem{cuadrado2023}
J.~Cuadrado, U.~Rançon, B.~R. Cottereau, F.~Barranco, and T.~Masquelier, ``Optical flow estimation from event-based cameras and spiking neural networks,'' \emph{Frontiers in Neuroscience}, vol.~17, p. 1160034, 2023.

\bibitem{li2021enhancing}
B.~Li, H.~Meng, Y.~Zhu, R.~Song, M.~Cui, G.~Chen, and K.~Huang, ``Enhancing 3-d lidar point clouds with event-based camera,'' \emph{IEEE Transactions on Instrumentation and Measurement}, vol.~70, pp. 1--12, 2021.

\bibitem{brebion2023learning}
V.~Brebion, J.~Moreau, and F.~Davoine, ``Learning to estimate two dense depths from lidar and event data,'' in \emph{Scandinavian Conference on Image Analysis}.\hskip 1em plus 0.5em minus 0.4em\relax Springer, 2023, pp. 517--533.

\bibitem{ronneberger2015u}
O.~Ronneberger, P.~Fischer, and T.~Brox, ``U-net: Convolutional networks for biomedical image segmentation,'' in \emph{International Conference on Medical image computing and computer-assisted intervention}.\hskip 1em plus 0.5em minus 0.4em\relax Springer, 2015, pp. 234--241.

\bibitem{dosovitskiy2017carla}
A.~Dosovitskiy, G.~Ros, F.~Codevilla, A.~Lopez, and V.~Koltun, ``Carla: An open urban driving simulator,'' in \emph{Conference on robot learning}.\hskip 1em plus 0.5em minus 0.4em\relax PMLR, 2017, pp. 1--16.

\bibitem{chen2020dynamic}
J.~Chen, J.~Meng, X.~Wang, and J.~Yuan, ``Dynamic graph cnn for event-camera based gesture recognition,'' in \emph{2020 IEEE International Symposium on Circuits and Systems (ISCAS)}.\hskip 1em plus 0.5em minus 0.4em\relax IEEE, 2020, pp. 1--5.

\bibitem{liu2021event}
Q.~Liu, D.~Xing, H.~Tang, D.~Ma, and G.~Pan, ``Event-based action recognition using motion information and spiking neural networks.'' in \emph{IJCAI}, 2021, pp. 1743--1749.

\bibitem{kim2016real}
H.~Kim, S.~Leutenegger, and A.~J. Davison, ``Real-time 3d reconstruction and 6-dof tracking with an event camera,'' in \emph{European conference on computer vision}.\hskip 1em plus 0.5em minus 0.4em\relax Springer, 2016, pp. 349--364.

\bibitem{guan2024evi}
W.~Guan, P.~Chen, H.~Zhao, Y.~Wang, and P.~Lu, ``Evi-sam: Robust, real-time, tightly-coupled event--visual--inertial state estimation and 3d dense mapping,'' \emph{Advanced Intelligent Systems}, vol.~6, no.~12, p. 2400243, 2024.

\bibitem{wu2024event}
X.~Wu, W.~He, M.~Yao, Z.~Zhang, Y.~Wang, B.~Xu, and G.~Li, ``Event-based depth prediction with deep spiking neural network,'' \emph{IEEE Transactions on Cognitive and Developmental Systems}, 2024.

\bibitem{zhu2018multivehicle}
A.~Z. Zhu, D.~Thakur, T.~{\"O}zaslan, B.~Pfrommer, V.~Kumar, and K.~Daniilidis, ``The multivehicle stereo event camera dataset: An event camera dataset for 3d perception,'' \emph{IEEE Robotics and Automation Letters}, vol.~3, no.~3, pp. 2032--2039, 2018.

\bibitem{shi2023even}
P.~Shi, J.~Peng, J.~Qiu, X.~Ju, F.~P.~W. Lo, and B.~Lo, ``Even: An event-based framework for monocular depth estimation at adverse night conditions,'' in \emph{2023 IEEE International Conference on Robotics and Biomimetics (ROBIO)}.\hskip 1em plus 0.5em minus 0.4em\relax IEEE, 2023, pp. 1--7.

\bibitem{tumpa2024snn}
S.~A. Tumpa, A.~Devulapally, M.~Brehove, E.~Kyubwa, and V.~Narayanan, ``Snn-ann hybrid networks for embedded multimodal monocular depth estimation,'' in \emph{2024 IEEE Computer Society Annual Symposium on VLSI (ISVLSI)}.\hskip 1em plus 0.5em minus 0.4em\relax IEEE, 2024, pp. 198--203.

\bibitem{Hidalgo20threedv}
D.~G. Javier Hidalgo-Carrio and D.~Scaramuzza, ``Learning monocular dense depth from events,'' \emph{{IEEE} International Conference on 3D Vision.(3DV)}, 2020.

\bibitem{velodyne}
Ouster, ``Vlp16: mid range lidar sensor,'' \url{https://ouster.com/products/hardware/vlp-16}, accessed:2025-09-14.

\bibitem{gallego2020event}
G.~Gallego, T.~Delbr{\"u}ck, G.~Orchard, C.~Bartolozzi, B.~Taba, A.~Censi, S.~Leutenegger, A.~J. Davison, J.~Conradt, K.~Daniilidis, \emph{et~al.}, ``Event-based vision: A survey,'' \emph{IEEE transactions on pattern analysis and machine intelligence}, vol.~44, pp. 154--180, 2020.

\bibitem{intelL515specs}
I.~Corporation, ``Intel\textsuperscript{\textregistered} realsense\texttrademark lidar camera l515,'' \url{https://www.intel.com/content/www/us/en/products/sku/201775/intel-realsense-lidar-camera-l515/specifications.html}, accessed: 2025-09-14.

\bibitem{prophesee}
M.~by~Prophesee, ``Evk4: The ultra high-speed and compact, hd event-based vision evaluation kit built to endure field testing conditions.'' \url{https://www.prophesee.ai/event-camera-evk4/}, accessed: 2025-09-14.

\bibitem{jetson}
NVIDIA, ``Nvidia jetson orin,'' \url{https://www.nvidia.com/en-sg/autonomous-machines/embedded-systems/jetson-orin/}, accessed: 2025-09-14.

\bibitem{papa2023meter}
L.~Papa, P.~Russo, and I.~Amerini, ``Meter: A mobile vision transformer architecture for monocular depth estimation,'' \emph{IEEE Transactions on Circuits and Systems for Video Technology}, vol.~33, no.~10, pp. 5882--5893, 2023.

\bibitem{godard2019digging}
C.~Godard, O.~Mac~Aodha, M.~Firman, and G.~J. Brostow, ``Digging into self-supervised monocular depth estimation,'' in \emph{Proceedings of the IEEE/CVF international conference on computer vision}, 2019, pp. 3828--3838.

\end{thebibliography}
